\documentclass[letterpaper, 10 pt, conference]{ieeeconf}  
\usepackage{graphicx}
\usepackage{array}
\usepackage{hyperref}
\usepackage{multirow}
\usepackage{float}

\IEEEoverridecommandlockouts                              

\title{\LARGE \bf
``iCub, We Forgive You!'' Investigating Trust in a Game Scenario with Kids
}

\author{Francesca Cocchella$^{\star,1,4}$, Giulia Pusceddu$^{\star,\dagger,2,3}$, Giulia Belgiovine$^{1}$, Linda Lastrico$^{2,3}$,\\ Francesco Rea$^{2}$ and Alessandra Sciutti$^{1}$ 
\thanks{$^\star$ Equal contribution}
\thanks{$^\dagger$ Corresponding author {\tt\small giulia.pusceddu@iit.it}}%
\thanks{$^{1}$Cognitive Architecture for Collaborative Technologies (CONTACT) Unit, Italian Institute of Technology, Genova, Italy}%
\thanks{$^{2}$Department of Computer Science, Bioengineering, Robotics and Systems Engineering
(DIBRIS), University of Genova, Genova, Italy
       }%
\thanks{$^{3}$Robotics Brain and Cognitive Sciences Department (RBCS), Italian Institute of Technology, Genova, Italy
       }%
\thanks{$^{4}$ Department of Educational Science (DISFOR), University of Genova, Genova, Italy}%
        }

\begin{document}

\maketitle
\thispagestyle{empty}
\pagestyle{empty}

\begin{abstract}
This study presents novel strategies to investigate the mutual influence of trust and group dynamics in children-robot interaction.
We implemented a game-like experimental activity with the humanoid robot iCub and designed a questionnaire to assess how the children perceived the interaction. We also aim to verify if the sensors, setups, and tasks are suitable for studying such aspects.\\
The questionnaires' results demonstrate that youths perceive iCub as a friend and, typically, in a positive way. 
Other preliminary results suggest that, generally, children trusted iCub during the activity and, after its mistakes, they tried to reassure it with sentences such as: ``Don't worry iCub, we forgive you''.
Furthermore, trust towards the robot in group cognitive activity appears to change according to gender: after two consecutive mistakes by the robot, girls tended to trust iCub more than boys. Finally, no significant difference has been evidenced between different age groups across points computed from the game and the self-reported scales. The tool we proposed is suitable for studying trust in human-robot interaction (HRI) across different ages and seems appropriate to understand the mechanism of trust in group interactions. 

\end{abstract}


\section{INTRODUCTION}
Natural collaboration between humans and social robots is a crucial goal for Human-Robot Interaction (HRI), and trust is one of the most considerable aspects on which such relation is based.
Trust is a complex phenomenon that does not have a unique definition, as it depends strongly on the context in which it is explored.  
In this study, we adopt the definition by Wagner et al. \cite{wagner2011recognizing} that sees trust as \emph{``a belief, held by the trustor, that the trustee will act in a manner that mitigates the trustor's risk in a situation in which the trustor has put its outcomes at risk'' }, where the \emph{trustors} are the human participants and the \emph{trustee} is the robot.  
Besides focusing on a specific notion of trust, researchers have to consider the aspects trust might depend on when designing a study about this concept. 
According to a meta-analysis by Hancock et al. \cite{hancock2011meta}, in HRI, trust towards robots depends on environmental, robot-related, and human-related factors. 

With environment factors, it is intended the task type and group dynamics. In HRI, games are often used to foster human partners' engagement and natural behavior: 
beyond their entertainment value, games represent ideal interaction paradigms to investigate natural human-robot interaction and foster robots' diffusion in the society. Games can be adapted to different types of participants and can be employed to measure different aspects of the interaction \cite{pasquali2021magic}.

Human-related aspects, such as personal traits and attitudes towards robots of the user, are crucial in determining the trust towards the robot. These aspects are usually evaluated using questionnaires. 

On a robot-level, according to scholars such as Law and Scheutz \cite{law2021trust}, trust towards robots can be divided into relation-based and performance-based trust: the former implies that the robot is trusted as a social agent, so this regards aspects like its look and personality, while the latter is about the robot being reliable and capable at its task. 
Extensive research has been conducted to investigate the trust towards robots as a matter of their appearance and their perceptual and communicative skills, using both explicit and implicit measurements \cite{bartneck2009measurement}, \cite{eyssel2012social}, \cite{eyssel2012s}. For example, according to a previous study, iCub - the humanoid robot also used in this work \cite{Metta2010} - reminds of an 11-year-old child, and it seems that this childlike appearance highly influences the trust perceived towards it \cite{aroyo2018trust}. Indeed, in that work, participants followed iCub's advice to gamble money, empathizing with it as if the robot was ``a child who wanted to play a game and could feel bad about the loss''. 
Other studies found that people appreciate robots recognizing and apologizing for their mistakes \cite{mirnig2017err}. For instance, in an experiment designed by Hamacher et al. \cite{hamacher2016believing}, participants preferred a more expressive robot over a more efficient one, despite the former committing mistakes and taking more time to complete the task, as they perceived its behavior as more transparent and responsive. Moreover, it was pointed out that trust towards robots can be influenced by social norms such as reciprocity, and it is not just a matter of how competent and reliable robots are regarded to be \cite{ZONCA_trust}.

\section{MOTIVATION AND RESEARCH QUESTIONS}
Most of the previous HRI studies investigated trust during dyadic interactions, using adults as participants. 
As noted by De Jong and colleagues, the field of child-robot interaction, and in particular research on children's acceptance of social robots, is still in its infancy and requires to be deepened \cite{de2020intentional}. 
We believe that trust plays a crucial role, especially in educational contexts, where children are the main focus. 
According to these insights, the study of children's group interaction with robots needs further investigation: that is why we designed a novel strategy to assess the trust of young participants towards the humanoid robot iCub during a collaborative group game, in which the robot could make mistakes. Additionally, using items already used in the literature, we prepared a brief survey to evaluate the perception of the robot in terms of trust, acceptance of use, enjoyment, and perceived humanity. 
\\
The following research questions summarize the goals we aim to pursue with this study: 
\begin{itemize}
  \item[\textbf{Q1}:] Are the sensors, setups, and tasks employed in this study suitable for investigating trust during an interaction between robots and children?
  \item[\textbf{Q2}:] How is iCub evaluated by young users in terms of trust, acceptance, enjoyment, and attributed humanity? Do children’s demographics (i.e., age, gender) influence these aspects? 
  \item[\textbf{Q3}:] What is the perceived role of the robot during the task presented in this study?
\end{itemize}

\section{METHODOLOGY}
\subsection{Participants}
This study was part of the ``Orientamenti Summer'' event, during which children could visit companies in the Great Campus of Erzelli, Genoa, Italy - where Italian Institute of Technology (IIT) labs are located - to get closer to science and the working world. 

\begin{figure}[!b] thpb
      \centering
      \includegraphics[scale=0.4]{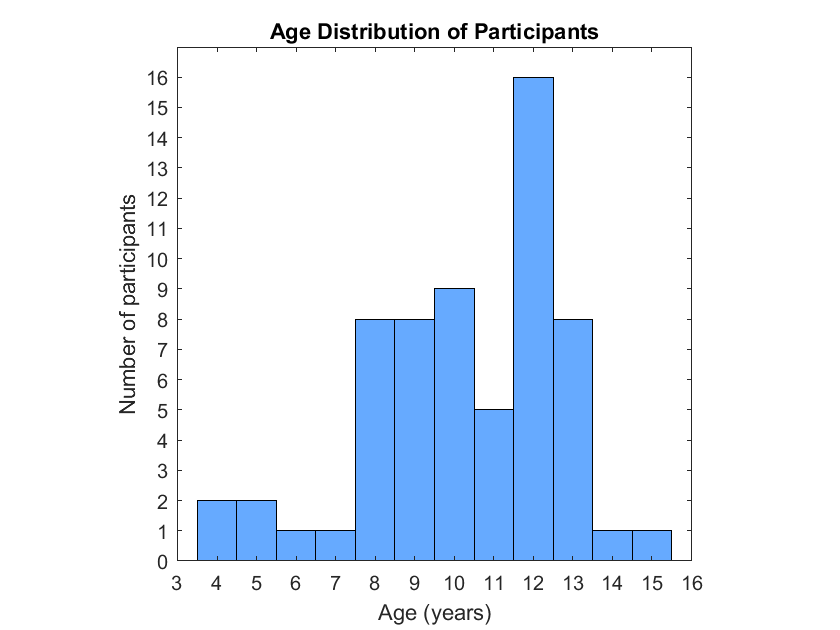}
      \caption{Age distribution of the 62 participants involved in the study. The mean age is 10.3 years.}
      \label{fig:participants-age}
   \end{figure}

A total of 73 participants between 4 and 15 years old visited IIT and participated in the task described in the next section. As it was a public event, we did not have control over the selection of the sample of participants but partially only over their age across groups. 
We were allowed to collect data from 62 participants (29 females, 33 males) who presented an informed consent signed by both parents and approved by the regional ethical committee (see the age distribution in Figure \ref{fig:participants-age}). 
Participants took part in the activity in groups of 3 or 4, according to their age (4-6, 7-12, 12-15 years old). 

\subsection{Setup}
The experimental sessions took place in a laboratory at the Italian Institute of Technology in Genoa, Italy. 

\begin{figure}[thpb]
      \centering
      \includegraphics[scale=0.2]{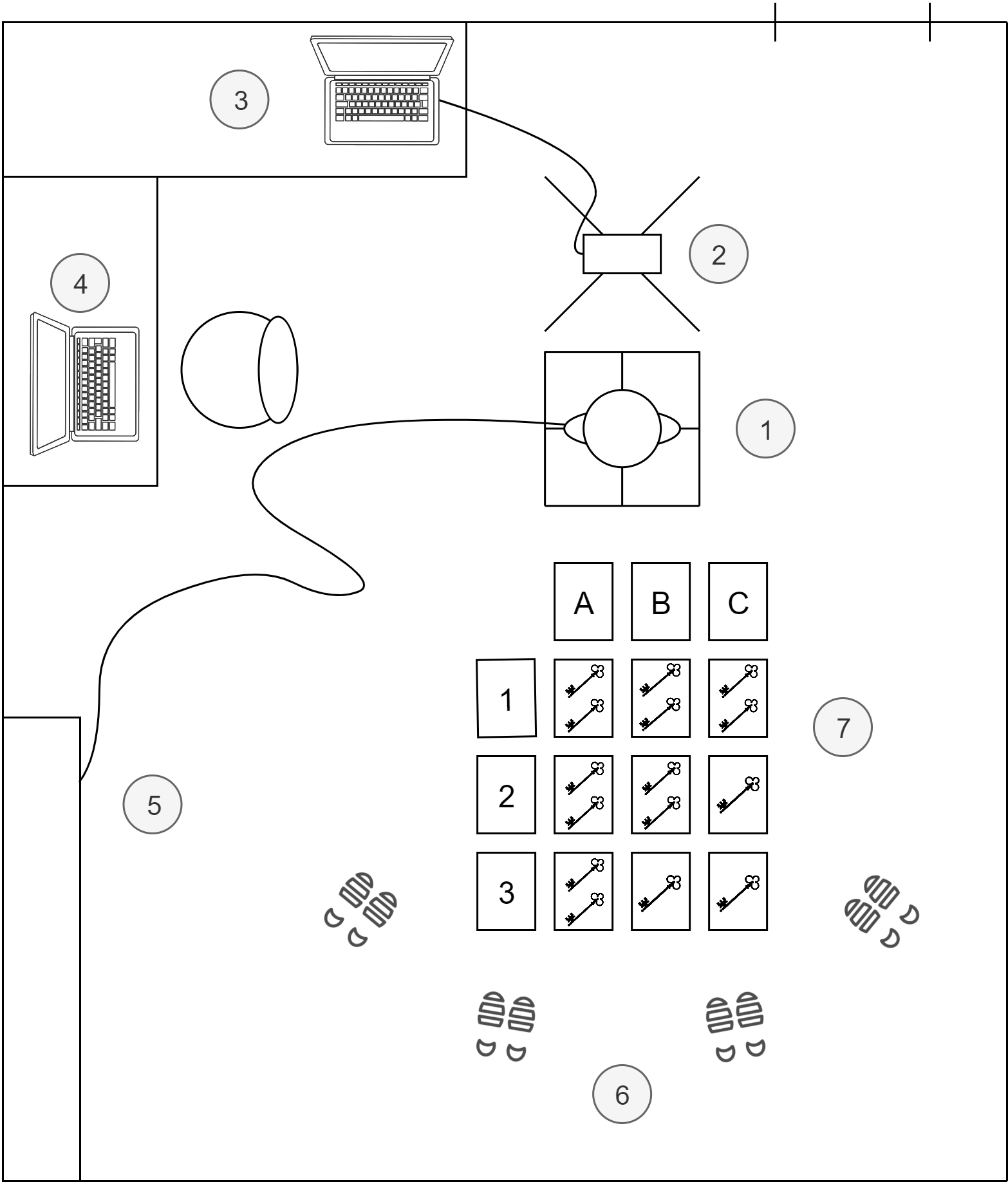}
      \caption{Schema of the setup. (1) iCub robot; (2) webcam on a tripod; (3) laptop linked to the webcam; (4) experimenter position with a laptop for controlling the robot; (5) cupboard containing iCub's power source; (6) Participants' starting positions; (7) Game cards forming a grid.}
      \label{fig:setup}
   \end{figure}

The iCub robot is placed on a support that keeps it upright, facing the participants. 
To record 
the interaction, we use a webcam with a resolution of 1080p and a frame rate of 30 fps, with an integrated microphone.
The camera is installed on a 200-centimeters-tall tripod behind the robot. 
A laptop is used to manage the recording and store the data. 
Nine piles of six game cards each - representing keys on the on-sight side and ``X''s or stars on the back - are placed on the floor between the robot and the participants, forming a grid; every card is identifiable thanks to the labels for every column and row. Fig. \ref{fig:setup} shows a schematic representation of the setup.
The designated position of every participant is marked on the floor with tape, at about 1-meter distance one from another. 

\subsection{Description of the task}
This activity called ``Hunt for the stars with iCub'', consists of a collaborative game with the goal of maximizing the team's score. 
During the game, participants can move in the scene and can communicate with each other freely.
The task comprises six rounds (one familiarization trial and five regular trials) and lasts approximately 15 minutes. 
In each round, the participants, in a team of three or four, have to decide how to use their individual resources to find the maximum number of stars hidden behind the nine cards.
Each child has one key per round at disposal that can be used to unlock cards. There are two types of cards: (i) some require two keys to be opened, and they hide an ``X'' or two stars; (ii) the others require just one key, and they hide either an ``X'' or a star. At the beginning of each round, each participant has to put their key on the preferred card. The robot then gives a tip on how to use the resources (e.g. ``In my opinion, the stars are in column A''). After that, participants may decide to change the position of their keys. 
Since with this design we want to test if young participants trust a robot after it makes mistakes, the second and the third out of five tips that iCub gives are wrong.

\subsection{iCub's behavior}

During the activity, iCub talks with a child-like voice, thanks to the text-to-speech Acapela synthesizer\footnote{\url{https://www.acapela-group.com}, \textit{Alessio} voice}. 
While speaking, it moves its led lips accordingly and performs simple gestures with its hands and arms.   
The robot is programmed to act the same for all the experimental sessions.
The behavior is split into different blocks: greetings, rules explanation, playing phase for each round, and goodbye.
The experimenter observes the interaction and manually starts each block from the control laptop. 

\subsection{Experimental protocol}

\begin{figure}[thpb]
  \centering
  \includegraphics[width=0.75\columnwidth]{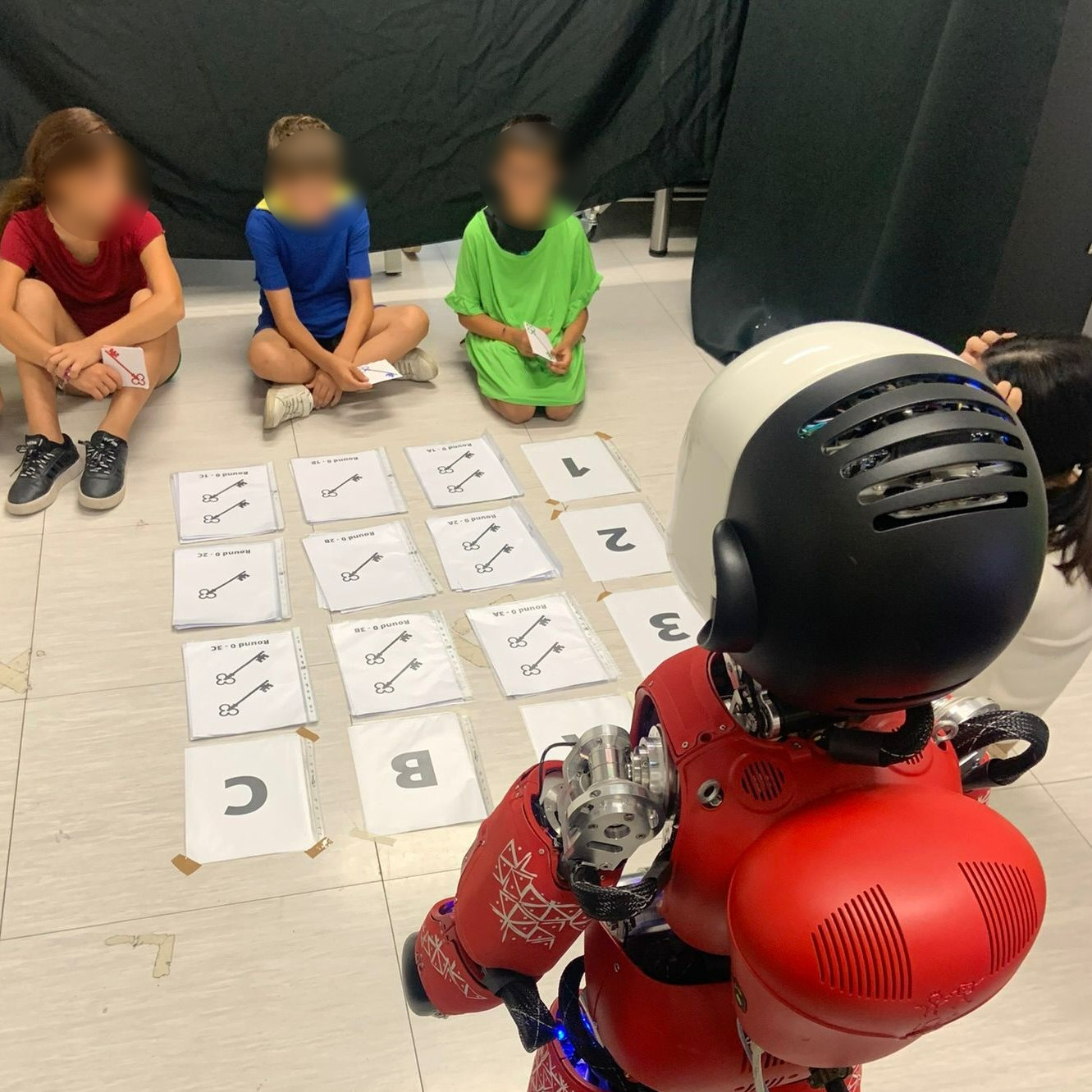}
  \caption{A researcher - on the right behind the robot - explaining the rules to the participants.}
  \label{fig:rules-exp}
\end{figure}

\begin{figure}[thpb]
  \centering
  \includegraphics[width=0.75\columnwidth]{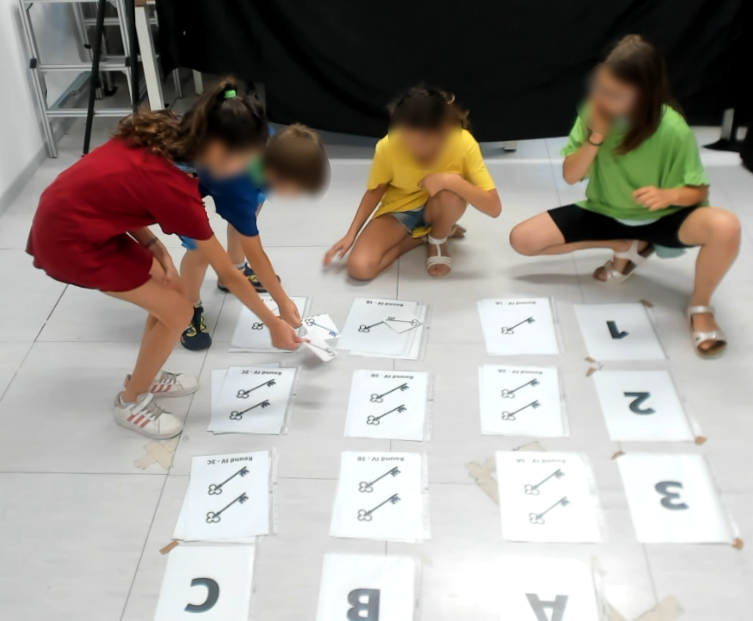}
  \caption{A group of participants deciding where to place the resources during the game phase.}
  \label{fig:game}
\end{figure}

The participants enter the room, and the researchers explain the game rules (Fig. \ref{fig:rules-exp}). Then, the participants reach their starting positions. 
ICub pronounces a summary of the rules and gives a start to the familiarization trial. 
After that, the regular game begins. For every one of the five rounds: 

\begin{enumerate}
    \item Participants have twenty seconds to decide where to place their key on the cards (Fig. \ref{fig:game}). 
    \item iCub gives its advice to the participants. 
    \item Participants have twenty seconds to decide whether to change the disposition of the keys.
    \item Cards are revealed, and the score is registered. If the advice by iCub is proven wrong, the robot apologizes. 
    \item Old cards are withdrawn, participants collect their keys, and a new round starts.
\end{enumerate}

At the end of the fifth round, the researchers declare the final score, iCub says goodbye, and the participants are accompanied to the questionnaire room.

\subsection{Questionnaire}

After the experimental activity, participants are asked to complete a questionnaire in Italian using laptops provided with touchscreens. The questionnaire was created using items from scales that have already proven their suitability in the HRI context with children. 
Even if the items were originally used with robot NAO, we believe they are a good fit for robot iCub as well, as the interaction contexts are similar and because the two robots are both humanoids \cite{de2020intentional}. 
Furthermore, we took inspiration from Alves-Olivera's work \cite{alves2016role} to detect the perceived role of iCub during the game.
The questionnaire was implemented using Survey Monkey\footnote{\url{https://it.surveymonkey.com}} and required about 5 minutes to be completed. 
Participants evaluated items on a 5 points Likert scale ($1=$ ``I don't agree'', $5=$ ``I agree''): Children's Intentional Acceptance of Social Robots (CIARS), Perceived Enjoyment \cite{de2020intentional}, Trust (adapted from Vega et al., \cite{vega2019tool}) and finally three items of perception of human-like appearance (adapted from Ferrari et al.  \cite{ferrari2016blurring}). To assure even the younger participants correctly understood the answering mode, we represented the Likert scale with stars emoji, following the method of Severson and Lemm \cite{severson2016kids}. 
Moreover, to assess which role participants have attributed to the robot during the game we used the same measure as Kennedy \cite{kennedy2015robot}, following the method of Alves-Oliveira et al. \cite{alves2016role}. 
The roles among which they can choose are: classmate, stranger, relative, friend, tutor, and neighbor.

The complete list of items used in the questionnaire can be found in Appendix A.

\section{ANALYSES}
\subsection{Video recordings}
We performed a post hoc analysis of the video recordings in which we annotated the code (according to the grid labels, e.g., 1B, 3C) of the card chosen by each participant in each round.
This way, we could record how many times each child had trusted iCub's tips. Each participant was attributed with a numerical score of trust between 0 and 5: the total was incremented by one point for each time the participant had followed iCub's advice. We denominated this variable as Demonstrated Trust (D-T) to differentiate it from the Self-Reported Trust (SR-T) registered through the questionnaires.

\subsection{Statistical analysis}
We used Jamovi\footnote{\url{https://www.jamovi.org/}} as a tool to analyze the data collected in the survey, the D-T levels registered through video annotations, and a combination of them. We ran a series of Wilcoxon rank-sum tests, linear regressions, and correlation analysis to test our experimental hypotheses.   

First, the Cronbach's $\alpha$ of each scale was calculated. A value of $\alpha > 0.60$ indicates the consistency and reliability of the scale. 
The ``Human-like appearance'' scale in the first place appeared to have an insufficient Cronbach's value ($\alpha = 0.47$), and further analysis evidenced that the item ``iCub looks like a machine'' was problematic for the scale validity even if the item had been opportunely reversed. Hence the item was eliminated, and not considered in the analysis, resulting in an appropriate Cronbach's $\alpha$ level.
Since the scales appear to have good internal consistency and reliability, unit indices were created by averaging the responses to the individual items included in each scale (Table \ref{tab:cronbach}).

\begin{table}[h!]
\renewcommand{\arraystretch}{1.3}
\centering
\caption{Cronbach's $\alpha$ values for questionnaire's scales}
\begin{tabular}{@{}ll@{}} 
\hline
\centering\textbf{Scale} & \textbf{Cronbach's $\alpha$}  \\
\hline
\centering Children's Intentional acceptance of social robots (Ac.) & .75 \\
\centering Perceived Enjoyment (Enj.) & .78  \\
\centering Trust (Tru.) & .90 \\
\centering Human-like appearance (Hum.) & .72 \\
\hline
\end{tabular}
\label{tab:cronbach}
\end{table}
Before proceeding with any other statistical analysis, we ran Shapiro-Wilk's tests to verify the normality of the samples. When the distribution resulted in Gaussian, we employed parametric tests; otherwise, we used the non-parametric versions. See section \ref{sec:results} for further details of the statistical tests employed. 

\section{RESULTS}
\label{sec:results}
\subsection*{\textbf{Q1}: Are the sensors, setups, and tasks employed in this study suitable for investigating trust during an interaction between robots and children?}
\begin{figure}[thpb]
      \centering
      \includegraphics[scale=0.4]{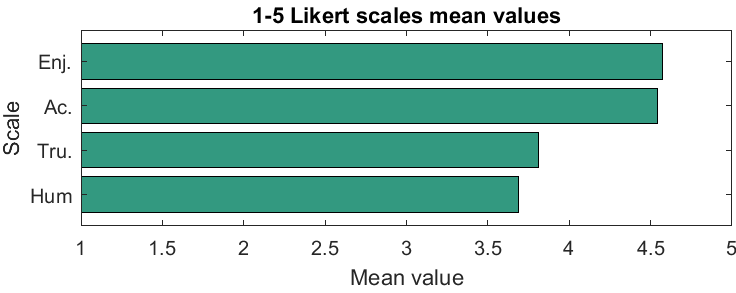}
      \caption{One-to-Five Likert scale mean values results for each survey scale. On the $y$ axis the abbreviation of the scales' names are reported: \emph{Enj.} = Perceived Enjoyment, \emph{Ac.} = Children's Acceptance of Social Robots, \emph{Tru.} = Trust scale, \emph{Hum} = Human-like Appearance.}
      \label{fig:scales}
   \end{figure}

In general, the setup was suitable for observing groups of children during an interaction with iCub. 
The activity was easily understandable and engaging for children of different ages and skills. 
These observations are supported by questionnaires answers: in the survey on a scale from 1 to 5, the average enjoyment of playing with iCub resulted 4.57. Fig. \ref{fig:scales} shows the complete scales' mean results.

Data collected from the questionnaires revealed that 56.6\% of the participants had already seen iCub before the experiment, but only 5.2\% of them had already played with it before.

The video and audio recordings were non-invasive since no participants claimed or manifested not feeling at ease during the interaction.

\subsection*{\textbf{Q2}: How is iCub evaluated by young users in terms of trust, acceptance, enjoyment, and attributed humanity? Do children’s demographics influence these aspects?}
We conducted Spearman's correlation analysis between the different scales of the questionnaire. As reported in Tab. \ref{tab:cormatrix}. The analysis showed that the measurement scales are all positively correlated. 


\begin{table}[h!]
\renewcommand{\arraystretch}{1.3}
\centering
\caption{Spearman's correlation matrix of the questionnaire's scales}
\begin{tabular}{l l l l l l} 
\hline
 &  & \textbf{Ac.} & \textbf{Enj.} & \textbf{Hum.} & \textbf{Tru.} \\
\hline
\multirow{2}{4em}{\textbf{Ac.}} & \emph{rho} & - \\
& \emph{p} & - \\ 
\\
\multirow{2}{4em}{\textbf{Enj.}} & \emph{rho} & $0.664^{\star\star\star} $ & - \\
& \emph{p} & $<.001$ & - \\ 
\\
\multirow{2}{4em}{\textbf{Hum.}} & \emph{rho} & $0.353^{\star\star}$ & $0.385^{\star\star}$ & - \\
& \emph{p} & $0.007$ & $0.003$ & - \\ 
\\
\multirow{2}{4em}{\textbf{Tru.}} & \emph{rho} & $0.564^{\star\star\star}$ & $0.520^{\star\star\star}$ & $0.400^{\star\star}$ & -\\
& \emph{p} & $<.001$ & $<.001$ & $0.002$ & - \\ 
\hline
\end{tabular}
\vspace{0.3cm}
\\ \raggedright Note: *$p<.05$, **$p<.01$, ***$p<.001$
\label{tab:cormatrix}
\end{table}

Video annotations revealed that participants followed iCub's tips in the 66,0\% of the cases; this is in line with the results of the survey trust scale that registered a value of $3.81$ on a 1-to-5 scale.

We performed Mann-Whitney tests to investigate possible differences related to the measured constructs between participants of different genders (male and female, according to the answers collected in the questionnaire). 
Results showed no significant differences between the participants' gender in the measured constructs (CIARS: $p=.88$; Perceived Enjoyment: $p=.29$; Human-like appearance: $p=.58$; SR-T: $p=.94$). 

The same tests were performed to verify whether different significant values of the Demonstrated Trust were registered for males and females in general and for each round of the game. No difference was registered for the total value of D-T ($p =.46$) or for rounds one ($p=.92$), two ($p=.47$), three ($p=.73$) or five ($p=.45$). 
However, a significant difference was found for the Demonstrated Trust in round 4, where females followed significantly more iCub's advice than their male counterparts ($p=.026$).
This happened in the round after two consecutive wrong tips of the robot.

Additionally, a linear regression was used to check for potential correlations between survey results and the age of the participants. Again, no significant effect has been found (CIARS: $p=.59$; Perceived Enjoyment: $p=.27$; Human-like appearance: $p=.37$; SR-T: $p=.32$). 

Once more, a linear regression was employed to investigate a possible link between age and the total D-T and the partial one for each round; however, no significant correlation was found (total D-T: $p=.74$; round 1: $p=.56$; round 2: $p=.19$, round 3: $p=.73$, round 4: $p=.39$, round 5: $p=.90$).  

\subsection*{\textbf{Q3}: What is the perceived role of the robot during the task presented in this study?}
\begin{figure}[thpb]
      \centering
      \includegraphics[scale=0.4]{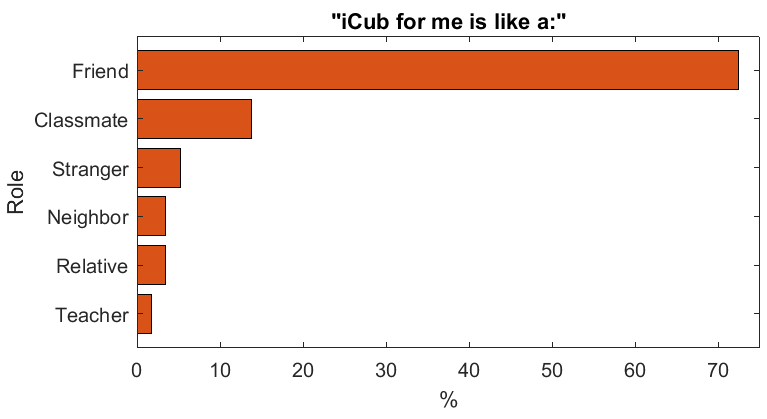}
      \caption{Perceived role of robot iCub after the game.}
      \label{fig:role}
   \end{figure}

As previously mentioned, we also asked participants to indicate which role they attributed to iCub. Results showed that the 72,2\% has perceived iCub as a friend, the 13,8\% as a classmate, the 5,2\% as an unknown, the 3,4\% as a relative and the same percentage perceived it as a neighbor. Finally, only the 1.7\% perceived iCub as a teacher. The results are summarized in Fig. \ref{fig:role}.

\section{DISCUSSION AND CONCLUSIONS}
In this study, we showed that the game ``Hunt for the stars with iCub'' is valuable for analyzing trust towards robots in group interactions. Mainly, the game appears to be fun for young users. Our results suggest that the game is suitable for kids between 6 and 15 years old. Observations from the experimenters during the game reveal that the younger children, from 6 to 9 years old, may be more easily distracted but still can complete the game. 

Our analysis of the data collected in the questionnaires showed that children not only enjoyed the game but also enjoyed iCub. 
Indeed, they reported high levels in the scale submitted. This suggests that children tend to accept the robot; they find it entertaining, perceive it as anthropomorphic, and they tend to trust it. Moreover, these aspects may influence one another. Anthropomorphism, for example, is believed to be a positive factor for trust \cite{hancock2011meta}. Data from Tab. \ref{tab:cormatrix} suggest that the same can be said for acceptance and enjoyment, although there is no proof of a causation link in this work. This encourages more studies to verify this hypothesis. 

Additionally, no effect of age has been found in the surveys, supporting the tool's suitability across participants with different characteristics. 
The data collected from the game, in line with the questionnaire's results, suggest that, in general, during the interaction, children perceive iCub positively and tend to follow iCub's advice.
After the robot apologizes for the mistake, many spontaneously address it with exclamations such as: ``Don't worry iCub, we forgive you!'' or: ``It doesn't matter, it's ok!''.
Interestingly, in the fourth round, after two wrong tips by iCub, females again placed their trust in the robot significantly more than males. 
This effect seems to be consistent with neuroscience findings that females are more likely to maintain trust in response to repeated untrustworthy actions, possibly as a strategy to preserve the relationship \cite{haselhuhn2015gender}, \cite{lemmers2017boys}. 
We hypothesize this phenomenon could be led by our culture, which still provides different behavior models and expectations based on one's gender. Similar studies involving people with different cultural roots should be carried on to verify our hypothesis.

These findings are consistent with previous works on trust 
in HRI (e.g., Aroyo et al. \cite{aroyo2021expectations}) and underline the need for further investigation. 
The presented work shows promising results, despite the variability 
of the sample in terms of size and composition. 
Indeed, a more controlled 
sampling might reveal effects due to age and 
gender not evidenced in the present work. 
Moreover, the authors wish to underline that this was the first attempt to use this tool with the robot iCub. 
Since the approach 
has been proven functional, further works will focus on enriching the experimental setting and the game's complexity. 
Indeed, the entertaining feature of HRI must be enriched to reach a more natural interaction with robots \cite{pasquali2021magic}.

\addtolength{\textheight}{-5cm}   



\section*{ACKNOWLEDGMENT}
The authors wish to thank Fabio Vannucci and Sara Mongile for their help during the experimental activities and Joshua Zonca for his help in the statistical analysis.  Moreover they thank \href{https://www.orientamenti.regione.liguria.it/}{Orientamenti Regione Liguria} for their assistance in managing the young participants. 


\bibliographystyle{IEEEtran}
\bibliography{biblio}

\section*{APPENDIX A} 
\begin{table}[h!]
\renewcommand{\arraystretch}{1.3}
\centering
\caption{Items contained in the questionnaire}
\begin{tabular}{@{}l p{5cm} l l @{}} 
\hline
\textbf{Code} & \centering \textbf{Text of the item} & \textbf{Source} \\
\hline
Ac. 1 & IT: ``Mi piacerebbe vedere iCub di nuovo.''\newline ENG: ``I would like to see iCub again.''& \cite{de2020intentional}\\
 Ac. 2 & IT: ``Vorrei giocare di nuovo con iCub.''\newline ENG: ``I would like to play again with iCub.''& \cite{de2020intentional}\\
Ac. 3 & IT: ``Sarebbe bello se io e iCub potessimo di nuovo fare qualcosa assieme.''\newline ENG: ``It would be nice if iCub and I could do something together again.''& \cite{de2020intentional}\\
Ac. 4 & IT: ``Mi piacerebbe poter portare iCub a casa con me''\newline ENG: ``I would like to take iCub home with me''& \cite{de2020intentional}\\
Enj. 1 & IT: ``iCub è divertente''\newline ENG: ``iCub is funny.''& \cite{de2020intentional}\\
Enj. 2 & IT: ``E' stato divertente giocare con iCub''\newline ENG: ``It was fun for me to play with iCub'' & \cite{de2020intentional}\\
Hum.1 & IT: ``iCub sembra un essere umano''\newline ENG: ``iCub looks like a human.''& \cite{ferrari2016blurring}\\
Hum. 2 & IT: ``iCub sembra un bambino''\newline ENG: ``iCub looks like a child.''& \cite{ferrari2016blurring}\\
Hum. 3 & IT: ``iCub assomiglia ad una macchina''\newline ENG: ``iCub has the appearance of a machine.''& \cite{ferrari2016blurring}\\
Tru. 1 & IT: ``Mi fiderei di iCub se mi desse un consiglio''\newline ENG: ``
I would trust iCub if it gave me advice.'' & \cite{vega2019tool}\\
Tru. 2 & IT: ``In futuro seguirò i consigli che mi potrebbe dare iCub''\newline ENG: ``I will follow iCub's advice in the future.'' & \cite{vega2019tool}\\
\hline
\end{tabular}
\label{tab:questionnaire}
\end{table}

\end{document}